\documentclass[11pt]{article}
\usepackage[a4paper,margin=1in]{geometry}
\usepackage[T1]{fontenc}
\usepackage[utf8]{inputenc}
\usepackage{lmodern}
\usepackage{microtype}
\usepackage{amsmath,amssymb}
\usepackage{graphicx}
\graphicspath{{figures/}}
\usepackage{booktabs}
\usepackage{xcolor}
\usepackage{hyperref}
\usepackage[numbers,sort&compress]{natbib}
\usepackage{caption}
\usepackage{enumitem}
\hypersetup{colorlinks=true,linkcolor=blue!50!black,citecolor=blue!50!black,urlcolor=blue!50!black}
\newcommand{\ci}[2]{[#1,\,#2]}
\newcommand{\thisdoi}{10.5281/zenodo.23163706}
\newcommand{\evidencedoi}{10.5281/zenodo.23163299}
\newcommand{\doilink}[1]{\href{https://doi.org/#1}{#1}}

\title{Moving the Mean Toward the Known Good, Not Beyond It:\\ What Inference-Time Interventions and Weight Consolidation\\ Buy in Open-Ended Generation}
\author{Roberto I. Ono Filho\\ Independent researcher\\
\texttt{ono.roberto@gmail.com} \quad ORCID \href{https://orcid.org/0009-0006-8650-629X}{0009-0006-8650-629X}}
\date{\small Version 2, 5 October 2026 (arXiv:2608.28886v2; Zenodo \doilink{\thisdoi}), with the corrections listed after the abstract.\\ Version 1: 28 August 2026 (arXiv:2608.28886v1; Zenodo \doilink{10.5281/zenodo.22714136}).}

\begin{document}
\maketitle

\begin{abstract}
What does a generation loop gain from learning on its own verified
successes? In cycles of generate, verify, select and LoRA-consolidate on
online bin packing, training on value-filtered candidates shifts what
the model writes on held-out variants toward value ($-1.7$ points of
excess, $p=0.008$; $-3.1$ against a random-consolidation control,
$p=0.004$) while the best observed candidate converges to the classic
heuristic's level and no further. A
confirmation battery replicates the whole procedure three times, with
fresh seeds and a never-consulted held-out set read exactly once: the
mean was nearly identical in all three lineages ($-2.0$, $-1.8$, $-1.9$),
and after aggregating within held-out variant all seven evaluable
variants favored consolidation ($p=0.008$). The best observed candidate
moved \emph{to} the classic heuristic's level, exactly (0.021028, the
lowest variant's classic level, in all three lineages, for attract and
for the random control alike); attract never went beyond its variant's
classic level, while the random control did in 7 of 24 lineage-variant
cells. A matched SFT-only control shows
the supervised anchor, not repulsion from bad candidates, does the
concentrating (96\% of candidates land exactly at the classic
heuristic's level). On far families the tails cut both ways: consolidation lowers the
per-candidate rate of better-than-classic candidates (10\% to 3.9\%)
while its larger production yields more such candidates absolutely (5
against 1, on few events). As motivation we report the inference-time
ledger that led here: a model-written schematic recap buys judged
document integration and nothing buys development; a verifier written
into the stream is imitated, 16.4 fabricated verdict lines per notebook.
In this arrangement (an 8-bit model in MLX, five cycles), mean quality
among valid candidates can be bought and replicated; under attraction the
observed best goes to the classic and, so far, never beyond it.
\end{abstract}

{\small\noindent\textbf{Changes in version 2.} An audit of version 1
against the run data and the notebook found errors, corrected here.
(i) The replication's ``no beyond-classic candidate anywhere'' was wrong
for the random control: 14 of its 529 valid candidates beat their
variant's classic level (7 of 24 lineage-variant cells, margins
0.0009--0.0037), as did 1 of the base's 147; attract's 432 did not
(abstract, Sections~\ref{sec:intro}, \ref{sec:related},
\ref{sec:consolidation}, \ref{sec:discussion}, \ref{sec:conclusion},
Table~\ref{tab:crep}). (ii) The SFT-only adapter's far-family mean was
the pooled value (0.1278); per family it is 0.0786, close to the anchored
arm's, so ``generalizes worst'' is withdrawn. (iii) Smaller statements
and numbers: strict finds before the final cycle (rare, not absent); the
captions of Table~\ref{tab:c} and Figure~\ref{fig:meanbest}; the
anchored arm as one of four with no far better-than-classic candidate;
per-variant and per-family values in three rows of
Table~\ref{tab:ladder}; the base's
far yield per notebook (0.06); three ranges; ``favoring attract'' for
negative differences; registration dates, with an SFT-only row
(Table~\ref{tab:provenance2}). (iv) Version 1 placed all run data and
judgments in the repository, but those of batteries M, M+, S and V were
not public; they are now in an evidence deposit, and the judging dates
are 18--20 August.
(v) Added: a limitation on a later bf16/PyTorch run, a statement on AI
assistance and on who read the drafts, and bibliography fixes.\par}

\section{Introduction}
\label{sec:intro}

\citet{ono2026interrupting} dismantled a cognitively inspired generation
loop over 24 conditions and found that most of its effect on judged novelty
lives in one operation, the \emph{interruption}: a new subject injected
every few hundred tokens into a stream whose literal repetition is damped.
On windows of fresh generated text the interruption raises judged surprise
by 1.2 to 1.4 points over habituation alone, on three base models, a
post-trained one and two genres, and a pre-registered replication confirms
it. The same study closed with three negatives. Read as whole documents,
none of the streams builds anything (no arm above 2.5 on integration,
development, coherence or surprise, 0--10, judged on 4,500 tokens with the
injected text removed); a judge-gated review that decides when not to
interrupt does not change that; and on online bin packing, where a
deterministic verifier replaces the judge, the interruption multiplies the
valid, distinct candidate heuristics a base model writes three- to fourfold
without raising the quality of the best. The interruption is a variation
operator, and variation without selection is organized noise.

This paper takes up the question that study left: what would make local
gains compose? We proceed by giving the loop, one piece at a time, the
things a mind that discovers has and the loop did not, and measuring each
under the protocol that the companion study showed to be necessary
(generated-only windows, fresh-only estimates, a document-level reading, the
premise as the unit of inference, a deterministic verifier where one
exists). The pieces are (i) a \emph{schematic} memory, in which the model
carries a compressed recap of its own stream instead of the verbatim text
that entrains it \citep{bartlett1932remembering}; (ii) a \emph{standing
question}, in which the interruption points into the stream's own
unresolved matter rather than away from it \citep{zeigarnik1927behalten};
(iii) \emph{selection}, a FunSearch-style evolution over the candidates the
interruption multiplies \citep{romeraparedes2024funsearch}; (iv) a
\emph{verifier inside the stream}, whose verdict on each candidate is
written back into the notebook; and (v) \emph{consolidation}, STaR/ReST
cycles in which the model is fine-tuned (LoRA) on its own
training-verifier-selected candidates
and generates again \citep{zelikman2022star,gulcehre2023rest,singh2023restem},
followed by preference-based repulsion \citep{rafailov2023dpo} and by
quality-diversity selection \citep{mouret2015mapelites,lehman2011novelty}
designed to keep the tails. The first two pieces are reported briefly as
the study's motivation; the paper's core is the consolidation thread and
its replication. Every battery was pre-registered before it ran;
the pre-registrations, their dated amendments and the results are in the
laboratory notebook released with the code.

\paragraph{Contributions.}
(1) Consolidation by LoRA on training-verifier-selected candidates moves
the candidate distribution on never-seen variants of the trained family
toward value ($p=0.008$; against a random-consolidation control,
$p=0.004$), moves the observed best to the classic heuristic's level and
not beyond it, and does not transfer to other families. (2) A confirmation battery: three independent
lineages with fresh seeds and a never-consulted held-out set read once
find the same mean shift in each ($-2.0$, $-1.8$, $-1.9$ points; all
seven evaluable variants favor consolidation, $p=0.008$) and the same
convergence of the best, exactly at the classic level in every lineage;
a matched SFT-only control shows the supervised anchor, not the
repulsion term, does the in-family concentrating, and out of family the
two anchored adapters are close (0.0786 vs 0.0758).
(3) The tails, both ways: on the far families the per-candidate
better-than-classic rate is highest in the untrained base while attraction's larger production
yields more absolute tail events; pass@$k$ against the classics is zero
for base, attract and random on the first held-out set at cycle five. (4) As motivation, the inference-time ledger: a model-written
schematic recap gives the program's highest document integration at no
local cost, a verifier written into the stream is imitated rather than
used (16.4 fabricated verdict lines per notebook), and no injection
moves development past the progression wall. (5) A reading of the whole:
inference-time stimulation reorganizes a fixed prior; weight
consolidation moves its mass toward the known good, with the per-candidate
cost in the tails and the production gain reported side by side.

\section{Related work}
\label{sec:related}

\paragraph{Self-training on one's own outputs.} STaR \citep{zelikman2022star}
fine-tunes a model on its own rationales that led to correct answers; ReST
\citep{gulcehre2023rest} and ReST-EM \citep{singh2023restem} alternate
sampling, filtering by a reward and supervised fine-tuning, and report gains
that saturate after a few iterations. Our attract arm is that recipe at
office scale (an 8B model, forty examples per cycle, a rank-8 adapter) with
three additions: a same-size random-consolidation control, which separates
``trained on its own outputs'' from ``trained on what the verifier
approved''; transfer measured on variants and families that never entered
any training set; and the rate of candidates better than the classic
heuristics, which is where the cost of self-training shows.

Our consolidation result is close to the finding that reward training
concentrates probability on existing trajectories, raising low-$k$
performance while large-$k$ coverage stays or falls
\citep{yue2025rlvr}. The literature also contains the counterexamples
that sharpen its scope: boundary-guided curriculum RL reports raising
both pass@1 and pass@256 above the base \citep{cai2026curriculum}, and
representation-based exploration bonuses improve pass@$k$ in
post-training \citep{tuyls2025representation}; both add what is absent
here, external guidance or directed exploration beyond the model's own
rollouts.

\paragraph{Preference optimization.} DPO \citep{rafailov2023dpo} optimizes a
policy against a frozen reference from chosen/rejected pairs. Our repulsion
arms use it with the model's own worst candidates, or the clones of the
known optimum, as rejected examples; unanchored it degenerates, anchored
with a supervised term on the chosen side it sharpens the policy onto the
known optimum. Both behaviours are known failure and success modes of
preference optimization; what we add is their effect on the tails of a
candidate distribution measured by a verifier.

\paragraph{Quality-diversity.} MAP-Elites \citep{mouret2015mapelites} and
novelty search \citep{lehman2011novelty,stanley2015greatness} keep the best
solution per behavioural niche rather than the best overall. Our qd arm
consolidates one elite per niche of the verifier's per-instance behaviour
profile; it keeps functional diversity and, in this regime, loses value and
the tails alike.

\paragraph{Verified search.} FunSearch \citep{romeraparedes2024funsearch}
and AlphaEvolve \citep{novikov2025alphaevolve} pair an untrained proposal
model with a hard evaluator and many samples, without fine-tuning the
proposer. Our results are consistent with that choice: on the far
families, the untrained prior has the highest rate of better-than-classic
candidates, and every consolidation we ran lowered it.

\paragraph{Memory, tension, and the judge.} Schematic memory as
reconstruction \citep{bartlett1932remembering}; unfinished tasks as
retained tension \citep{zeigarnik1927behalten}; compression progress as
intrinsic reward \citep{schmidhuber2010formal}. On the instrument, the
companion study documents what a windowed LLM judge cannot see (injected
text, self-copy beyond its horizon, windows against wholes); the fabricated
verdicts of Section~\ref{sec:selection} are the same entrainment acting on a
verifier's output placed in the prompt.

\section{Setup}
\label{sec:setup}

\paragraph{Generators and loop.} Unless stated, the generator is
Qwen3-30B-A3B-Base (mixture of experts, 8-bit, MLX on an Apple M5 Pro);
battery C uses Qwen3-8B-Base (8-bit) because it is fine-tuned. A cell is
one premise (the ten premises of the companion study) continued for 4,500
tokens with the end-of-text token masked; habituation (windowed repetition
penalty, window 512, factor 1.15) is on in every arm; interruptions fire on
a clock every 300 tokens. Arms differ in what is injected and in whether
the context is kept or rebuilt.

\paragraph{Judging.} Windows of generated text only (96 tokens, starting 32
after any injected text, never crossing an injection), judged by Claude
Opus~5 ($k=5$, median) on surprise, connection and coherence with 600 tokens
of context; windows flagged as self-copy (half their 12-token shingles seen
earlier in the stream) are excluded from the fresh-only estimates. The whole
stream, injected text removed, is judged once per cell ($k=3$) on
integration, development, coherence and surprise. The premise is the unit:
cell means, bootstrap intervals over cells, exact sign-flip permutation
tests on paired cells, one-sided where pre-registered.

\paragraph{The problem with a verifier.} Online bin packing in the
FunSearch form: the model writes a notebook of \texttt{priority(item,
remaining)} functions; the verifier packs instances and reports the mean
excess over the lower bound. Batteries S and V use the ten distribution
variants of the companion study (item ranges within $[0.1, 0.8]$, where
best fit is near-optimal); battery C moves to a family of small-item
distributions in which best fit is measurably beatable
(Section~\ref{sec:consolidation}).

\section{Memory and tension: the inference-time motivation}
\label{sec:memory}

This section is the study's motivation, kept brief; the paper's core is
the consolidation thread that follows. Five arms varied what an
interrupted stream remembers and what the interruption asks (period 300,
ten premises): \emph{verbatim} keeps the full context across
interruptions; \emph{reset} rebuilds it from the premise and the new
subject; \emph{schema} rebuilds it from the premise, a two- to
three-sentence recap the model writes about its own last 1{,}200 tokens,
and the new subject; \emph{question} keeps the context and injects the
open question the model extracts from its own stream; \emph{agenda}
combines reset, recap and question. Recaps and questions count as
injected text (never judged; removed from the document), with an
anti-copy ladder for the model's tendency to repeat its own recap.

\begin{table}[t]
\centering\small
\begin{tabular}{@{}lcccc@{}}
\toprule
arm (Qwen3-30B-A3B-Base, period 300) & integration & development & coherence & surprise \\
\midrule
habituation, no interruption & 1.40 & 1.10 & 1.50 & 1.40 \\
verbatim context (clock300) & 0.80 & 0.40 & 1.50 & 0.50 \\
reset + new subject & 2.00 & 1.60 & 2.20 & 2.50 \\
schema: reset + recap + new subject & \textbf{2.70} & 1.80 & 2.30 & 2.70 \\
question: preserved + open question & 1.50 & 1.10 & 1.90 & 1.00 \\
agenda: reset + recap + open question & 2.50 & \textbf{2.00} & 1.90 & 2.60 \\
\bottomrule
\end{tabular}
\caption{Document-level reading (whole 4{,}500-token stream, injected text
removed; Opus~5, $k=3$; cell means over ten premises, 0--10). The schematic
recap gives the highest integration of the program; no arm passes 2.0 on
development.}
\label{tab:doc}
\end{table}

Pre-registered contrasts (one-sided exact paired permutation, ten
premises): schema beats verbatim on the composite by $+1.65$
\ci{+1.15}{+2.25}, $p=0.001$; against reset, the registered composite is
not supported ($+0.45$, $p=0.13$; decomposed, integration $+0.70$
\ci{+0.10}{+1.30}, development $+0.20$), so most of the headline is the
reset and the recap adds integration on top, at no cost on fresh windows
(all fresh contrasts n.s., self-copy 0\%). The question beats the
neutral subject under preserved context ($+0.70$, $p=0.004$) but matches
habituation on development; the agenda adds nothing over schema
($+0.20$, $p=0.38$; the document judge: ``restarts the premise over and
over with new casts and genres''). The reading that motivates the rest
of the paper: remembering the schema beats re-reading the text, but
every registered injection moves integration and never development above
about 2/10, the \emph{progression wall}.

\section{Selection and the verifier}
\label{sec:selection}

\paragraph{Battery S: the interruption inside a selection loop.} A
FunSearch-style evolution (10 variants $\times$ 8 generations $\times$ 16
samples per generation; the prompt shows the current population; paired by
variant) with and without the interruption's angle line injected into the
prompt. S1 (gain of the champion on held-out instances): $+0.0013$
\ci{-0.0015}{+0.0047}, $p=0.38$. S2 (cumulative distinct valid candidates):
89.5 vs 95.1, $p=0.86$ against the hypothesis. S3 (generation of first
escape from best fit): $+0.5$, n.s. Under selection the interruption adds
nothing, not even the diversity it added in a single stream: the population
prompt already supplies what the interruption supplied.

\paragraph{Battery V: the verifier inside the stream.} The notebook of the
companion study's battery B, generated in chunks; whenever a
\texttt{priority} function closes, the harness scores it on the training
instances and writes the verdict into the notebook as a comment
(``\texttt{\# Verifier: priority\_x scored mean excess 0.0812 on the
training instances (worse than best fit, 0.0752); no improvement.}''). A
second arm adds, every 300 tokens, a deterministic scoreboard (``\texttt{\#
So far: N functions tried; the best is \ldots; best fit sits at \ldots}'')
and a value-anchored open question. Ten variants $\times$ two seeds;
baselines are battery B's plain and angle arms re-scored in text order.
V1 (held-out gain, feedback vs angle): $+0.0003$, $p=0.48$. V2 (candidates
improving on the notebook's own training best, the verifier's measure of
progression): 0.1 vs 0.3 per notebook, in the opposite direction. V4
(scoreboard adds to feedback): nothing; the scoreboard arm writes fewer
functions than every other arm (1.1 per notebook). Progression is near zero in every arm
(0.1--0.3 improvements per notebook), with the standing caveat that best fit
is near-optimal on these distributions.

\paragraph{Fabricated verdicts.} The model wrote on average 16.4
``\texttt{\# Verifier:}'' lines per notebook in the feedback arm (6.2 with
the scoreboard) against about three real ones: verdicts with invented
scores and ``new best of this notebook'' that the harness never issued.
Injecting the \emph{format} of a reward signal into the context makes a
base model continue a highly predictable textual pattern present in its
context; whether the fabrication functions as reward or as mere format
imitation is untested (no fabricated verdict was produced by the harness;
their correlation with true scores and their effect on subsequent
candidates are open analyses). The rule is narrower than ``keep verdicts
out of the prompt'': do not place forgeable verifier syntax in the same
generative channel without provenance (a non-generable channel, a nonce,
reserved tokens) and without measuring imitation.

\section{Consolidation: moving the prior}
\label{sec:consolidation}

\paragraph{Design (pre-registered 2026-08-21, extensions 2026-08-21 and 2026-08-22).}
Qwen3-8B-Base (8-bit), QLoRA (rank 8, 16 layers, lr $10^{-5}$, batch 2,
prompt masked, $\approx$4 epochs over the consolidation set). Domain: a
family of small-item bin-packing distributions in which best fit is
beatable by simple heuristics (headroom 0.3--0.5 points of excess; first fit
ties or beats best fit), ten training variants and eight held-out variants
that never enter any training set; six far families (Weibull, triangular,
bimodal, OR-library) for far transfer; verification on $20\times200$-item
instances; a \emph{find} beats the better of best fit and first fit by 0.001
on the variant's training instances. Cycle 0 is a shared base generation (3
notebooks $\times$ 1,200 tokens per variant, an angle comment every 250
tokens). Cycles 1--5: \emph{attract} trains a fresh adapter on the finds
plus the top-40 candidates by training excess from its own lineage;
\emph{random} trains on a same-size random sample of valid candidates (the
control that separates ``trained on itself'' from ``trained on what is
worth''); \emph{base} generates without an adapter at the same seeds. On the
cycle-5 lineage we then trained: \emph{repel} (DPO, $\beta=0.1$, 80 steps,
chosen = the attract set, rejected = the worst valid candidate of the same
variant), \emph{repel-anch} (the same plus a token-mean supervised term on
the chosen side, $\alpha=1$), \emph{repel-early} (plain DPO, 20 steps),
\emph{qd} (supervised, one elite per behavioural niche of the verifier's
per-instance profile, $k=3$ niches per variant plus one seat for the best
clone of the classic attractor) and \emph{repel-mode} (anchored DPO whose
rejected examples are the best-fit clones themselves). Hypotheses, on the
held-out variants: C1 (primary), attract $<$ base on the mean test excess of
valid candidates (the prior moved on variants it never saw); C2, attract
$<$ random; C3, higher find rate; C4 (two-sided), the diversity cost; R1'
repulsion adds to attraction; R2' the best moves; T1 the tail rate (share of
candidates better than the classics) on far families; T2 functional
collapse.

\begin{table}[t]
\centering\small
\begin{tabular}{@{}lccccccc@{}}
\toprule
arm & cycle & candidates & mean excess & rel.\ to min(BF,FF) & best & distinct & copies \\
\midrule
base & 0 & 65 & 0.0633 & +0.0322 & 0.0408 & 0.99 & -- \\
base & 1 & 53 & 0.0562 & +0.0251 & 0.0327 & 1.00 & -- \\
attract & 1 & 235 & 0.0559 & +0.0248 & 0.0319 & 0.98 & 10\% \\
random & 1 & 276 & 0.0708 & +0.0397 & 0.0332 & 1.00 & 5\% \\
base & 2 & 42 & 0.0583 & +0.0272 & 0.0317 & 0.97 & -- \\
attract & 2 & 198 & \textbf{0.0430} & +0.0119 & 0.0312 & 0.99 & 6\% \\
random & 2 & 274 & 0.0697 & +0.0386 & 0.0310 & 0.99 & 8\% \\
base & 3 & 55 & 0.0644 & +0.0333 & 0.0414 & 1.00 & -- \\
attract & 3 & 214 & \textbf{0.0516} & +0.0204 & 0.0311 & 0.99 & 7\% \\
random & 3 & 334 & 0.0653 & +0.0342 & 0.0312 & 0.99 & 1\% \\
base & 4 & 57 & 0.0532 & +0.0221 & 0.0323 & 1.00 & -- \\
attract & 4 & 195 & \textbf{0.0420} & +0.0109 & 0.0311 & 0.98 & 16\% \\
random & 4 & 276 & 0.0510 & +0.0199 & 0.0307 & 1.00 & 0\% \\
base & 5 & 47 & 0.0556 & +0.0244 & 0.0325 & 1.00 & -- \\
attract & 5 & 226 & \textbf{0.0386} & +0.0074 & 0.0311 & 1.00 & 1\% \\
random & 5 & 287 & 0.0695 & +0.0383 & 0.0311 & 0.99 & 0\% \\
\bottomrule
\end{tabular}
\caption{Battery C on the eight held-out variants (never in any training
set; test instances): candidates per arm and cycle, mean test excess of
valid candidates (cell means over variants), the same relative to the
better of best fit and first fit, the best candidate, the distinct/valid
ratio and the share of literal copies of the arm's training set. The
value-filtered adapter lowers the mean; its best reaches the classic
level and stays there.}
\label{tab:c}
\end{table}

\begin{figure}[t]
\centering
\includegraphics[width=0.98\linewidth]{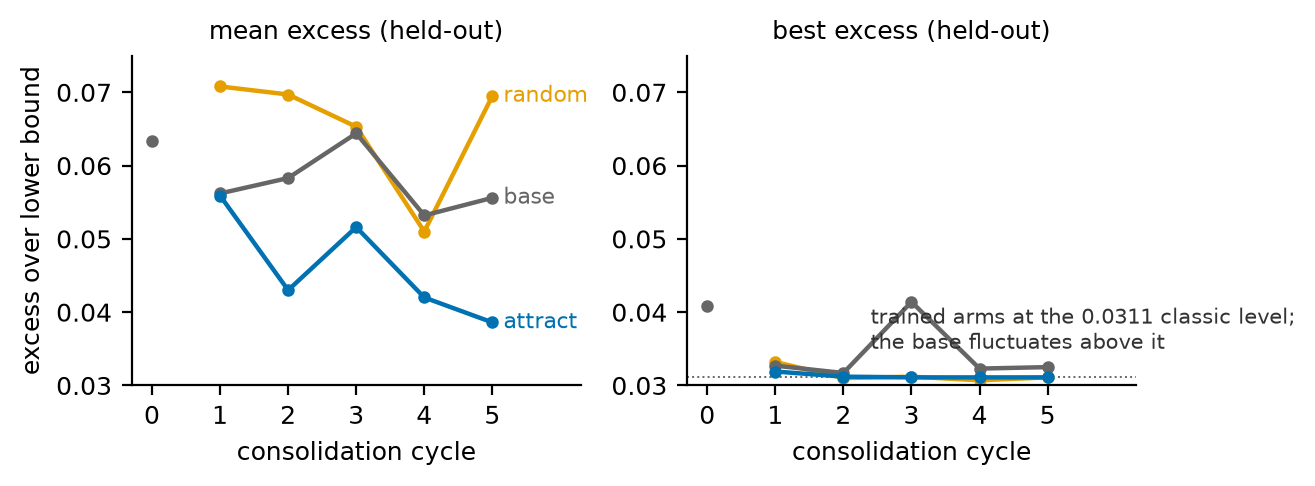}
\caption{The title of this paper, as data (held-out variants, never in any
training set; cycle 0 is the shared pre-consolidation base). Left: mean
excess over the lower bound falls under attraction and nowhere else.
Right: from the second cycle the trained arms' best stays within 0.0005
of the 0.0311 classic level, while the base's best fluctuates above it
(0.0317--0.0414). Consolidation moves the mean; attraction reaches but
does not exceed the classic level, while the random control dips below
it at cycles 2 and 4.}
\label{fig:meanbest}
\end{figure}

\paragraph{Attraction.} Table~\ref{tab:c} gives the held-out reading by arm
and cycle (Figure~\ref{fig:meanbest}). At the pre-registered final cycle (3), C1 was $-0.0129$
\ci{-0.0317}{+0.0027}, $p=0.109$, not supported at $\alpha=0.05$, with the
direction consistent (7 of 8 variants better) and the cycle-2 contrast
supported ($-0.0153$, $p=0.016$); a pre-registered extension to five cycles
closed it: $-0.0170$ \ci{-0.0268}{-0.0079}, $p=0.008$, the attract arm
ending at 3.86\% mean excess, 0.74 points above the better classic
heuristic, against 5.3--6.4\% for the base. C2: attract vs random,
$-0.0138$ ($p=0.035$) at cycle 3 and $-0.0309$ ($p=0.004$) at cycle 5: the
value filter matters. The random control hurts at cycle 1 ($+1.5$ points
over base: consolidating mediocre functions makes the next ones worse) and
is neutral to harmful later. C3: strict finds are rare (at most 1.7\%,
attract at cycle 4) and zero in every arm at the final cycle, and the
best candidate per variant is nearly the same in every arm
($\approx$3.1\% excess, the level of best fit): the ceiling did not move.
C4: no collapse over five cycles; the distinct/valid ratio is 0.97--1.00
throughout, literal copies of the training set 1--16\% (attract); both
adapters make the model write 3.4--6.5$\times$ as many valid functions
per notebook as the base.
Far transfer: on six distributions of other families the final attract
adapter equals the base on the mean ($+0.001$, $p=0.69$; $n=5$ of the six
far families; the base arm produced no valid candidate in one) and beats
the random adapter by $-0.020$ ($p=0.016$); the random adapter is worse
than the base there too ($+0.017$). What moved is the prior for the family
it was trained on. The adapter of the last cycle was taught mostly
tight-bin functions (24 of 40 of the form \texttt{1/(r**2+eps)}, 13
best-fit-like): consolidation pulls the mass toward the best attractor the
verifier has approved.

\begin{table}[t]
\centering\footnotesize
\setlength{\tabcolsep}{4.5pt}
\begin{tabular}{@{}lcccccc@{}}
\toprule
 & \multicolumn{4}{c}{held-out variants (8)} & \multicolumn{2}{c}{far families (6)} \\
\cmidrule(lr){2-5}\cmidrule(lr){6-7}
adapter (cycle-5 lineage) & cand. & mean exc. & at classic & levels & mean exc. & $>$ classics \\
\midrule
base (none) & 47 & 0.0556 & 20\% & 0.81 & 0.0838 & 10.0\% \\
random & 287 & 0.0695 & 16\% & -- & 0.1004 & 0.0\% \\
repel-early (DPO, 20 steps) & 71 & 0.0721 & 34\% & -- & 0.0994 & 2.4\% \\
attract (SFT on value) & 226 & 0.0386 & 72\% & 0.14 & 0.0803 & 3.9\% \\
repel-anch (DPO + anchor) & 144 & \textbf{0.0311} & 100\% & 0.07 & \textbf{0.0758} & 0.0\% \\
sft-only (anchor term alone) & 78 & 0.0319 & 97\% & -- & 0.0786 & 0.0\% \\
qd (SFT on niche elites) & 293 & 0.0598 & 25\% & 0.33 & 0.1012 & 0.6\% \\
repel-mode (DPO, from attractor) & 62 & 0.0792 & 2\% & 0.50 & 0.0935 & 0.0\% \\
repel (DPO, unanchored) & 13 & 0.1056 & 0\% & -- & -- & -- \\
\bottomrule
\end{tabular}
\caption{The ladder of adapters. Held-out: mean test excess of valid
candidates, share scoring exactly at the level of the better classic
heuristic, distinct score levels per valid candidate; far families: mean
excess and share of candidates better than the classics. Means and shares
are averaged over variants (held-out) or families (far), as in
Table~\ref{tab:c}; candidate counts are totals. The mass walks
onto the known attractor; attraction lowers the far-family tail rate per
valid candidate while raising absolute tail yield through production.
Rows below attract are single-lineage mechanistic follow-ups.}
\label{tab:ladder}
\end{table}

\paragraph{Repulsion.} Unanchored DPO (loss 0.69 $\to$ 0.0015) degenerated:
1.1 closed functions per notebook (attract 9, base 2), 60\% comment lines
and recursive non-code; the few valid candidates were far worse than the
base (10.6\% vs 5.6\% excess). Pushing the policy away from bad functions,
unanchored, pushes it away from writing functions. Anchored (the same loss
plus a token-mean supervised term on the chosen side; pre-registered), the
repulsion writes again (144 valid candidates, $+12$ per cell over the base,
$p=0.02$) and adds to attraction: mean held-out excess 3.11\% against
3.86\% ($-0.0074$ \ci{-0.0104}{-0.0047}, $p=0.008$), which is exactly the
level of the better classic heuristic. The best candidate does not move
($0.0311$ in both arms), and the reason is the finding: \textbf{every}
held-out candidate of the anchored-repulsion adapter scores exactly at the
best-fit level (all are monotone ``fullest bin'' variants), against 72\%
for attraction, 16--20\% for the base and the random control, 34\% for an
early-stopped plain DPO (otherwise worse than attraction, $+0.034$,
$p=0.008$) and 0\% for the degenerate one (Table~\ref{tab:ladder}).
Hash-level diversity survives (1.00); functional diversity does not. On the
far families the anchored adapter is the best arm on the mean (7.6\% vs
8.4\% base) because ``be best fit'' is good everywhere, and, with the
random control, the SFT-only anchor and mode repulsion, one of four arms
with no candidate better than the classics (Table~\ref{tab:ladder}).

\begin{figure}[t]
\centering
\includegraphics[width=0.98\linewidth]{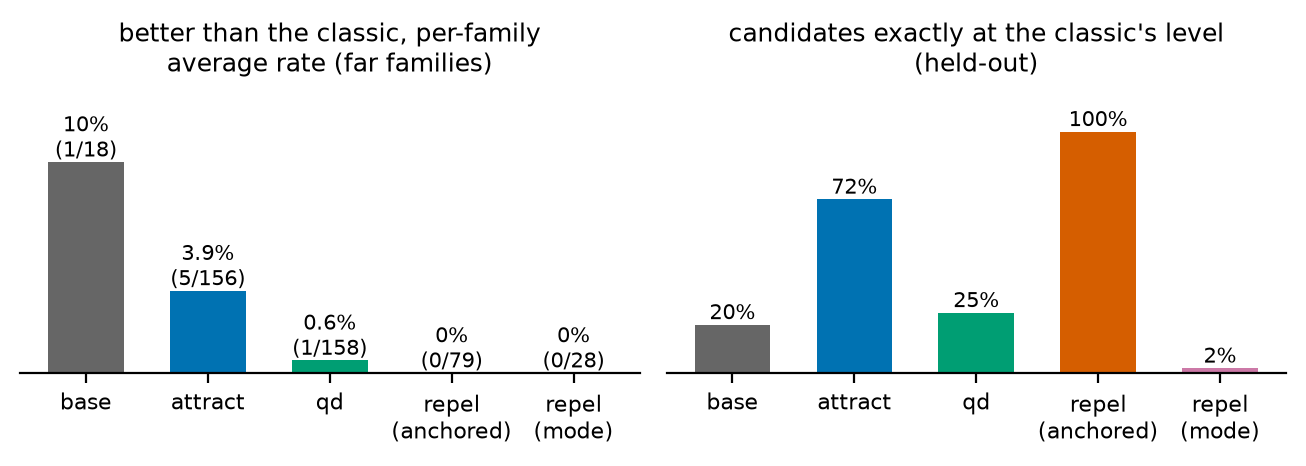}
\caption{What consolidation costs. Left: the share of candidates
\emph{better} than the stronger classic heuristic on the far families,
the tail where discovery lives, shrinks from 10\% (base) to 3.9\%
(attraction) to at most 0.6\% under every tail-preserving objective we
built. Right: on held-out variants the consolidated arms concentrate
\emph{exactly at} the classic's level (attraction 72\%, anchored
repulsion 100\%, a functional collapse), against 20\% for the base.
The bar heights are per-family average rates (the registered estimand);
the counts above them are pooled events over all candidates, a different
aggregation (base: 10.0\% macro but 1/18 = 5.6\% pooled; attraction:
3.9\% macro, 5/156 = 3.2\% pooled). Both rest on 0--5 events per arm
and are preliminary. Per notebook, attraction's absolute tail yield is
higher than the base's (0.28 vs 0.06). Competence per candidate is
bought with idiosyncrasy; total yield depends on production.}
\label{fig:tails}
\end{figure}

\paragraph{Equal-budget comparisons.} At matched candidate counts the
picture sharpens rather than reverses. Best-of-$k$ over 2{,}000
subsamples per variant: at $k{=}10$ valid candidates, attraction's
subsample best (0.0311) beats the random control's (0.0381) in eight of
eight variants each, while the base reaches ten valid candidates in only
one variant of eight (0.0449 there); at $k{=}40$ only the random arm has
coverage at all. The base's small-sample best advantage is therefore in
part a production deficit, and no arm's pass@$k$ against the classics
ever leaves zero. The pooled quantiles say the same about the whole
distribution: attraction dominates the base at p10, p50 and p90
(0.0257/0.0283/0.0676 against 0.0285/0.0483/0.1075), so the shift is of
the distribution, not only of its mean.

\paragraph{A two-part reading.} Because the headline measure conditions
on validity, the effect decomposes into two parts that the replication
battery lets us separate: consolidation multiplies the production of
valid candidates (base 33--63 per lineage against attraction's
108--184, with one base variant producing none at all) and, conditional
on validity, shifts their mean quality ($-0.019$). ``Moving the mean''
is the second part; the first may matter as much in practice, and both
replicate.

\paragraph{The registered SFT-only control (a single-lineage mechanistic
follow-up, like the repulsion, QD and far-family results that follow).}
The anchored objective
combines repulsion from the worst candidates with a supervised term on
the chosen ones; which half concentrates the policy? The matched control
(same pairs, same 80 steps, same learning rate and data order, the DPO
term removed) answers: the supervised anchor alone puts 75 of its 78
held-out candidates (96\%; 97\% as a mean over variants,
Table~\ref{tab:ladder}) exactly at the classic level (anchored: 100\%;
attraction: 72\%), at mean excess 0.0319 against the anchored arm's
0.0311. Repelling the bad was never the mover; supervised attraction to
the chosen does nearly all the concentrating, and the repulsion term
adds the last few points. On the far families the SFT-only adapter
averages 0.0786 (per-family mean; 0.1278 pooled, dominated by one
family), close to the anchored arm's 0.0758 and below the base's 0.0838,
with no better-than-classic candidate: the supervised anchor drives the
concentration in and out of the family; the DPO term adds little out of
it.

\begin{figure}[t]
\centering
\includegraphics[width=0.98\linewidth]{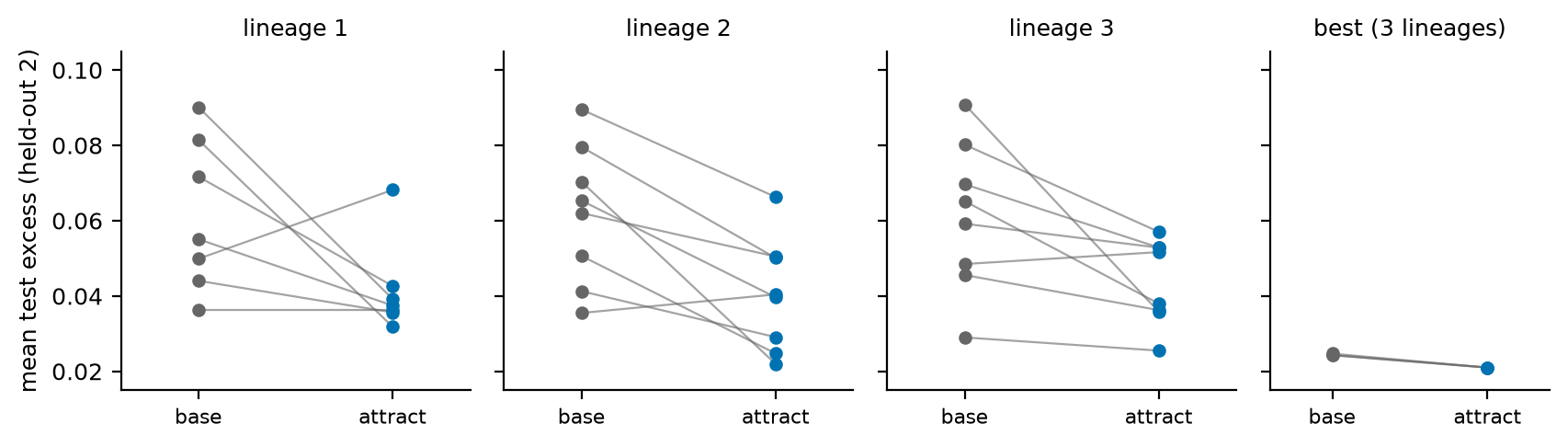}
\caption{Independent replication, variant by variant: each line pairs
one held-out variant's mean under base and under attract, one panel per
lineage; the right panel shows the best candidate. The slope repeats
across lineages; the best moves to the classic level in all three
lineages and stops there.}
\label{fig:crep}
\end{figure}

\paragraph{Independent replication across three lineages.} The
confirmation battery re-ran the whole procedure three times: independent
lineages with fresh generation and selection seeds, five cycles fixed in
advance, and a second held-out set of eight variants, disjoint from
every earlier set, generated and read exactly once after cycle five.
Table~\ref{tab:crep} and Figure~\ref{fig:crep} give the design's
answer. The mean moves in every lineage by nearly the same amount
($-0.0196$, $-0.0176$, $-0.0188$; pooled $-0.0187$
\ci{-0.0308}{-0.0080}, aggregated by held-out variant, favoring attract
in all seven evaluable variants, exact sign-flip $p=0.008$, hierarchical
bootstrap over lineages and variants); attract beats the random control
in each ($-0.0149$ pooled, $p=0.016$); and the best converges in all: the
adapters' best lands exactly at the classic level (0.021028, that of the
lowest variant) in each lineage. Per variant, attract never passes its
variant's classic level, while the random control does with 14
candidates (7 of 24 lineage-variant cells, margins 0.0009--0.0037) and
the base with one. Production replicates
too: the base wrote 33--63 valid candidates per lineage against the
adapters' 108--205, and in one variant of one lineage the base produced
no valid candidate at all (the eighth variant is therefore excluded from
the paired contrast, a conditioning we flag rather than hide). The
registered shared-pool one-shot (top-40 vs random-40 from a single base
pool) points the same way ($-0.008$, $p=0.078$ at $n{=}8$, underpowered
alone). Two readings sharpen the picture. First, a sensitivity for the
excluded variant: restoring it under imputation (assigning the base its
best or its worst observed variant mean there) gives $n{=}8$ with all
eight variants favoring attract, $p=0.0039$ under either bound, so the
conditioning was conservative. Second, a three-effect decomposition: the
random control also reaches the best of 0.021028 in every lineage, so
generic self-training plus increased valid production suffices to
\emph{reach} the classic level (and, in seven lineage-variant cells, to
pass it by at most 0.004); value filtering is what the distributional
mean shift requires.

\begin{table}[t]
\centering\footnotesize
\setlength{\tabcolsep}{4.5pt}
\begin{tabular}{@{}lcccccc@{}}
\toprule
lineage & arm & candidates & valid & mean excess & best & $\Delta$ vs base \\
\midrule
1 & base & 71 & 63 & 0.0612 & 0.0247 & \\
1 & attract & 201 & 184 & 0.0416 & 0.0210 & $-0.0196$ \\
1 & random & 219 & 191 & 0.0619 & 0.0210 & $+0.0007$ \\
2 & base & 79 & 51 & 0.0605 & 0.0243 & \\
2 & attract & 162 & 140 & 0.0430 & 0.0210 & $-0.0176$ \\
2 & random & 232 & 205 & 0.0548 & 0.0210 & $-0.0057$ \\
3 & base & 41 & 33 & 0.0613 & 0.0243 & \\
3 & attract & 120 & 108 & 0.0424 & 0.0210 & $-0.0188$ \\
3 & random & 202 & 133 & 0.0549 & 0.0210 & $-0.0064$ \\
\bottomrule
\end{tabular}
\caption{The replication battery on the never-consulted held-out set
(means averaged over the seven common variants; $\Delta$ = per-lineage
variant-averaged difference to base; best = minimum over all held-out
variants). Three independent runs of the whole procedure produce three
nearly identical effects on the mean and the same movement of the best:
to the classic level, in every arm that trains, and never beyond it
under attract.}
\label{tab:crep}
\end{table}

\paragraph{Keeping the tails.} Two objectives built to keep the tails
(pre-registered) recovered functional diversity and nothing else
(Figure~\ref{fig:tails}). The qd
adapter has 25\% of candidates at the classic level (attract 72\%,
$p=0.004$) and 2.4$\times$ the distinct score levels, at a cost of $+2.1$
points of mean excess ($p=0.008$), and its better-than-classic rate on the
far families is \emph{lower} than attraction's (0.6\% vs 3.9\%; T1, primary,
not supported). Repulsion from the attractor itself removes the collapse
entirely (2\% at the classic level) and with it most of the value (7.9\%
mean excess, best 4.8\%, worse than the base; no far tail). The diversity
these objectives keep is diversity of the mediocre. Across every
consolidation we ran, the per-valid-candidate rate of better-than-classic
candidates on new families is highest in the untrained base (10\% as a
mean over families; pooled, 1 of 18 candidates) and falls with training
(attract 3.9\%, pooled 5 of 156; every other adapter $\le 2.4\%$, pooled
0--1 events). The absolute yield points the other way: because
consolidation multiplies valid production, attraction delivered five
better-than-classic candidates against the base's one, 0.28 against 0.06
per notebook, across two families against one. Both readings rest on zero
to five events per arm and are preliminary; what they jointly say is that
consolidation concentrates each candidate toward the known good while
producing many more candidates, so the per-candidate tail rate falls even
where the total tail yield does not. The anchored objectives, by
contrast, produced no better-than-classic candidate at all.

\section{Discussion}
\label{sec:discussion}

\paragraph{What can be bought.} Across the two studies the ledger is now
complete for the interventions we could build. Variation is cheap: the
interruption raises fresh surprise by more than a point and multiplies valid
candidates three- to fourfold. The document judge assigned higher
integration to a schematic recap carried across resets than to any other
arm, although the pre-registered composite against reset alone was not
supported. Mass can be bought with the right training: a few cycles of
consolidation on what the verifier approved move the candidate distribution
of a never-seen variant onto the known good, and an anchored repulsion from
the bad moves it there exactly. Two things nothing we tried produced:
measurable document progression (three injections, selection, a verifier
in the stream and a judge-gated review all leave development near 2/10)
and, under the evaluated bin-packing budgets, a held-out candidate
measurably beyond the classic from any value-filtered or anchored adapter
(five cycles of attraction, four kinds of repulsion and a
quality-diversity consolidation leave the best observed held-out level
at the classic or above it, none passing it by more than $10^{-4}$; the
held-out candidates beyond it, by 0.0009--0.0037, came from the
untrained base and the random control).

\paragraph{The bit distance (a conceptual model consistent with, not
demonstrated by, these data).} A generative model is a distribution, not a
store; the question is not whether an answer is ``inside'' it but how much
probability mass sits near it, how many bits of search separate the prior
from the target. Inference-time stimulation reorganizes the surface of a
fixed prior: it buys variation and, with a schematic memory, integration,
but it does not move the mass. Weight consolidation moves the mass, and
battery C shows its present reach: toward the best thing the verifier had
already approved, within the family it was trained on, and at the cost of
the tails where the rare better-than-classic candidates appear; objectives
built to keep the tails kept diversity of the mediocre instead. This is
also why the known successes of verified search \citep{romeraparedes2024funsearch,
novikov2025alphaevolve} pair an untrained proposal distribution with a hard
evaluator and many samples rather than fine-tune the proposer. For discovery
our results argue for a portfolio, the broad untrained prior to propose and
the consolidated one to exploit, and for spending on selection.

\paragraph{Design rules.} For long open-ended generation: damp repetition;
interrupt with something new each time or reset the context; carry a
compressed recap rather than the text; do not expect the whole to advance
from any of this. For search with a verifier: spend on sampling and
selection, not on stimulation; keep the verifier's verdicts out of the
prompt or measure the fabrication they induce; if reliability is the goal,
consolidate on what the verifier approved (and never without the
filter); whether a portfolio that keeps the untrained prior in the loop
beats pure consolidation for discovery depends on the cost function
(per candidate, the base finds more; per notebook, attraction does) and
awaits a fixed-budget comparison.

\section{Limitations}
\label{sec:limitations}

Ten premises and ten variants per battery; 8--30B base models at 8 bits;
the document judge is a single LLM family at $k=3$ and its absolute levels
are low for every arm, so ``integration 2.7'' is a relative statement; the
bin-packing distributions of batteries S and V have little headroom, which
bounds what selection and feedback could show there (battery C's family was
chosen for headroom for that reason, and its headroom is still small:
0.3--0.5 points); the oracle that writes recaps and questions is the
generator itself and sometimes repeats itself, which the anti-copy ladder
mitigates but does not remove; the original battery C is one training
lineage of five cycles of forty examples on an 8B model with a rank-8
adapter, whose eight held-out variants replicate the adapter, not the
procedure; the replication battery (three independent lineages, new
generation and selection seeds, five cycles fixed in advance, a second
held-out set generated and read once) closes that gap for the mean and
the value filter, on seven of eight registered variants (the base
produced no valid candidate in one, the production deficit again); the
original lineage's primary
contrast missed $\alpha$ at the registered third cycle ($p=0.11$) and
closed only in a sequential-analysis sense at the extension to five
($p=0.008$), the held-out set having been consulted at both looks; its
base arm varies by seed; candidate counts differ by arm (42--334), so the
observed best is not an equal-budget comparison of tails (the matched-$k$
analyses in the results address this); the far families
are six (paired $n=5$) and the tail rates rest on tens of candidates; the
QD arm is a small implementation (three niches per variant), so
``tail-preserving objectives keep the mediocre'' is a statement about this
implementation, not the class \citep{wan2025loongflow}; after the
first cycle attract and random are two whole adaptive
procedures rather than two selection criteria applied to one pool (the
registered shared-pool one-shot control, top-40 vs random-40 from a
single base pool, points the registered way, $-0.008$, $p=0.078$ at
$n{=}8$, underpowered alone); excluding the far
family in which the base produced no valid candidate treats a production
failure as missing data, so the paired far contrast is conditioned on
validity; the selection criterion admits lucky candidates (training finds lose on
test), which is why the primary measure is the mean and not the best. A
later run (laboratory notebook, 2--3 September 2026; the appendices
\texttt{APPENDIX\_T2.md} and \texttt{APPENDIX\_T2\_8B.md} are in the evidence
deposit) repeated the consolidation on the same domain in
bf16 on PyTorch with three cycles (Qwen3-14B-Base and Qwen3-8B-Base) and
reproduced neither the base's far-family tail rate (0--5\% per cycle,
against 10\% here) nor attraction's mean shift on the held-out variants
(attract minus base between $-0.9$ and $+0.6$ points per cycle); the
claim that mean quality can be bought and replicated holds for the 8-bit
MLX arrangement with five cycles, and we have not separated which of
backend, precision, model size or number of cycles it depends on. The
negative results are resolution-bounded nulls, not proofs of absence; the
tail finding in particular deserves replication at a larger sample and on a
domain with more headroom.

\section{Conclusion}
\label{sec:conclusion}

Given memory, tension, selection and a verifier, the interrupted loop
composes better and advances no further. A schematic recap raised judged
integration relative to reset, while the pre-registered composite
contrast was not supported; a standing question outperformed the fixed
subject change under preserved context but did not improve development
over habituation alone or over schema; selection
and in-stream verification showed no detectable benefit on the
near-saturated variants tested, and the latter teaches the model to
imitate verdicts. Consolidating training-verifier-selected candidates
into the weights is the one intervention that moved what the model writes
on problems it had not seen, and it moved it toward what it already knew
to be good; the combined anchored objective moved it all the way there
and no further, and the registered SFT-only control showed the
supervised half does nearly all of that moving (96\% at the classic
level without any repulsion term);
keeping the tails kept the mediocre. In this lineage the untrained base
had the highest per-candidate better-than-classic rate on the evaluated
far families, while attraction's larger production delivered more
absolute tail events; the best statistic improved to the classic level,
and attraction produced no candidate measurably beyond the classic on
the held-out variants. In this arrangement (an 8-bit model in MLX, five
cycles), mean candidate quality among valid candidates can be bought;
under attraction the observed best moves to the classic heuristic and
not beyond it, in every lineage. And the central remaining question, whether
the procedure replicates or the effect was a peculiarity of one training
path, is now answered: three independent lineages, fresh seeds, five
cycles fixed in advance, a second held-out set read exactly once, and
the same result three times, $-2.0$, $-1.8$ and $-1.9$ points of mean
excess against the base (aggregated by held-out variant, favoring
attract in all seven evaluable variants, $p=0.008$), and the same
convergence of the best: in every lineage the adapters' best candidate
lands exactly on the classic heuristic's level (0.021028), and attract
produces no beyond-classic candidate (the random control 14, the base
one). The best statistic itself improves from the
base's 0.0243--0.0247 to that level, an attractor reading rather than an
immobility one; the exact sign-flip on the variant-paired best
difference does not resolve it ($-0.004$, $p=0.094$; the hierarchical
bootstrap interval \ci{-0.011}{-0.000} reflects a different aggregation
and should be read as descriptive).

\paragraph{Acknowledgements and tooling.} The experiments, analyses and
text of this paper were produced by the author working with an AI
assistant (Claude, Anthropic; model Fable 5 for version 1, August 2026,
and Opus 5.5 for the corrections of version 2, October 2026) inside the
Claude Code environment. The assistant wrote and ran the code, operated
the data collection under the author's direction and drafted the
manuscript; every design decision, result and claim was reviewed by the
author, who is responsible for the content. A colleague read the drafts
of version 1 in five rounds between 24 and 28 August 2026, and the
responses to each round are dated in the laboratory notebook; this was a
colleague's reading, not peer review. The corrections of version 2 come
from a read-only audit of version 1 against the run data and the
notebook, carried out by instances of the same assistant and approved by
the author. The LLM judge (Claude Opus~5 on Amazon Bedrock) is part of
the instrument; generators and fine-tuning ran locally on Apple silicon
via MLX.

\appendix
\section{Reproducibility and experimental provenance}
\label{app:repro}

\begin{table}[h]
\centering\footnotesize
\setlength{\tabcolsep}{4pt}
\begin{tabular}{@{}lllll@{}}
\toprule
battery & registered & data seen & primary & status \\
\midrule
S (selection) & 2026-08-19 & paper-1 batteries & held-out gain & null \\
M/M+ (memory, tension) & 2026-08-20 & paper-1, S & schema vs verbatim & supported \\
V (verifier) & 2026-08-20 & M, S & held-out gain & null; fabr.\ verdicts \\
C cycles 1--3 & 2026-08-21 & none & attract vs base (c3) & not supported ($p=0.11$) \\
C cycles 4--5 & 2026-08-21 & c3 result & attract vs base (c5) & sequential, $p=0.008$ \\
C-repel, N (tails) & 2026-08-21/22 & C & keep the tails & degenerate / not kept \\
Matched-$k$ analyses & 2026-08-24 & all C & best-of-$k$, pass@$k$ & reported \\
C-rep (3 lineages) & 2026-08-26 & all above & replication & mean yes; best to classic \\
SFT-only control & 2026-08-27 & all above & at-classic share & the anchor concentrates \\
\bottomrule
\end{tabular}
\caption{Order of registrations and looks (dates from the commit history;
M+ was registered after the M result). The cycle-5 result is a
prospective sequential extension, not an independent confirmation; the
independent lineages with never-consulted held-outs are the confirmation
battery.}
\label{tab:provenance2}
\end{table}

\begin{sloppypar}
Code, pre-registrations and the dated laboratory notebook are in the
repository of the companion study
(\url{https://github.com/RobertoOno/interrupting-the-loop}): the oracle arms
(\texttt{schema\_reseed}, \texttt{anomaly\_reseed}, \texttt{agenda\_reseed}
in \texttt{scripts/dream\_run.py}), battery S
(\texttt{scripts/evolve\_interrupt.py}), battery V
(\texttt{scripts/problem\_loop.py}), battery C and its extensions
(\texttt{scripts/consolidate.py}, \texttt{dpo\_lora.py},
\texttt{run\_c.sh}, \texttt{run\_c\_ext.sh}, \texttt{run\_c\_repel*.sh},
\texttt{run\_n.sh}) and the analyses that generate the tables
(\texttt{docs/APPENDIX\_GEN.md}, \texttt{APPENDIX\_S.md},
\texttt{APPENDIX\_V.md}, \texttt{APPENDIX\_C*.md}, \texttt{APPENDIX\_N.md}).
The candidate records of batteries C and C-rep are in the release
\texttt{run-data-v1} of that repository (Zenodo
\doilink{10.5281/zenodo.22714138}); the run data and judgments of every
battery reported here, including M, M+, S and V, are in the evidence
deposit \doilink{\evidencedoi}.
Judging used \texttt{anthropic.claude-opus-5} on Amazon Bedrock (18--20
August 2026) under the prompts of the companion study; fine-tuning used
\texttt{mlx\_lm} (QLoRA on the 8-bit model, peak 14\,GB, about 8 minutes
per adapter) and a small DPO trainer in MLX released with the code. Every
hypothesis, amendment and result is dated in \texttt{docs/PLANO.md}.
\end{sloppypar}

\bibliographystyle{plainnat}
\bibliography{references}

\begin{thebibliography}{17}
\providecommand{\natexlab}[1]{#1}
\providecommand{\url}[1]{\texttt{#1}}
\expandafter\ifx\csname urlstyle\endcsname\relax
  \providecommand{\doi}[1]{doi: #1}\else
  \providecommand{\doi}{doi: \begingroup \urlstyle{rm}\Url}\fi

\bibitem[Bartlett(1932)]{bartlett1932remembering}
Frederic~C. Bartlett.
\newblock \emph{Remembering: A Study in Experimental and Social Psychology}.
\newblock Cambridge University Press, 1932.

\bibitem[Cai et~al.(2026)Cai, Fang, Li, Zeng, Li, and Chen]{cai2026curriculum}
Pengxiang Cai, Tianchen Fang, Xiaohan Li, Qingyuan Zeng, Guocong Li, and Jintai
  Chen.
\newblock Curriculum reinforcement learning can incentivize reasoning capacity
  in {LLMs} beyond the base model, 2026.
\newblock arXiv:2606.22317.

\bibitem[Gulcehre et~al.(2023)Gulcehre, Paine, Srinivasan, Konyushkova, Weerts,
  Sharma, Siddhant, Ahern, Wang, Gu, Macherey, Doucet, Firat, and
  de~Freitas]{gulcehre2023rest}
Caglar Gulcehre, Tom~Le Paine, Srivatsan Srinivasan, Ksenia Konyushkova, Lotte
  Weerts, Abhishek Sharma, Aditya Siddhant, Alex Ahern, Miaosen Wang, Chenjie
  Gu, Wolfgang Macherey, Arnaud Doucet, Orhan Firat, and Nando de~Freitas.
\newblock Reinforced self-training ({ReST}) for language modeling, 2023.
\newblock arXiv:2308.08998.

\bibitem[Lehman and Stanley(2011)]{lehman2011novelty}
Joel Lehman and Kenneth~O. Stanley.
\newblock Abandoning objectives: Evolution through the search for novelty
  alone.
\newblock \emph{Evolutionary Computation}, 19\penalty0 (2):\penalty0 189--223,
  2011.

\bibitem[Mouret and Clune(2015)]{mouret2015mapelites}
Jean-Baptiste Mouret and Jeff Clune.
\newblock Illuminating search spaces by mapping elites, 2015.
\newblock arXiv:1504.04909.

\bibitem[Novikov et~al.(2025)]{novikov2025alphaevolve}
Alexander Novikov et~al.
\newblock {AlphaEvolve}: A coding agent for scientific and algorithmic
  discovery, 2025.
\newblock Google DeepMind; arXiv:2506.13131.

\bibitem[Ono~Filho(2026)]{ono2026interrupting}
Roberto~I. Ono~Filho.
\newblock Interrupting the loop: Periodic subject changes raise judged surprise
  and connection in base language models.
\newblock arXiv:2608.19893 [cs.CL], 2026.

\bibitem[Rafailov et~al.(2023)Rafailov, Sharma, Mitchell, Manning, Ermon, and
  Finn]{rafailov2023dpo}
Rafael Rafailov, Archit Sharma, Eric Mitchell, Christopher~D. Manning, Stefano
  Ermon, and Chelsea Finn.
\newblock Direct preference optimization: Your language model is secretly a
  reward model.
\newblock In \emph{Advances in Neural Information Processing Systems}, 2023.

\bibitem[Romera-Paredes et~al.(2024)Romera-Paredes, Barekatain, Novikov, Balog,
  Kumar, Dupont, Ruiz, Ellenberg, Wang, Fawzi, Kohli, and
  Fawzi]{romeraparedes2024funsearch}
Bernardino Romera-Paredes, Mohammadamin Barekatain, Alexander Novikov, Matej
  Balog, M.~Pawan Kumar, Emilien Dupont, Francisco J.~R. Ruiz, Jordan~S.
  Ellenberg, Pengming Wang, Omar Fawzi, Pushmeet Kohli, and Alhussein Fawzi.
\newblock Mathematical discoveries from program search with large language
  models.
\newblock \emph{Nature}, 625\penalty0 (7995):\penalty0 468--475, 2024.
\newblock \doi{10.1038/s41586-023-06924-6}.

\bibitem[Schmidhuber(2010)]{schmidhuber2010formal}
J{\"u}rgen Schmidhuber.
\newblock Formal theory of creativity, fun, and intrinsic motivation
  (1990--2010).
\newblock \emph{IEEE Transactions on Autonomous Mental Development}, 2\penalty0
  (3):\penalty0 230--247, 2010.

\bibitem[Singh et~al.(2023)Singh, Co-Reyes, Agarwal, Anand, Patil,
  et~al.]{singh2023restem}
Avi Singh, John~D. Co-Reyes, Rishabh Agarwal, Ankesh Anand, Piyush Patil,
  et~al.
\newblock Beyond human data: Scaling self-training for problem-solving with
  language models, 2023.
\newblock arXiv:2312.06585.

\bibitem[Stanley and Lehman(2015)]{stanley2015greatness}
Kenneth~O. Stanley and Joel Lehman.
\newblock \emph{Why Greatness Cannot Be Planned: The Myth of the Objective}.
\newblock Springer, 2015.

\bibitem[Tuyls et~al.(2025)Tuyls, Foster, Krishnamurthy, and
  Ash]{tuyls2025representation}
Jens Tuyls, Dylan~J. Foster, Akshay Krishnamurthy, and Jordan~T. Ash.
\newblock Representation-based exploration for language models: From test-time
  to post-training, 2025.
\newblock arXiv:2510.11686.

\bibitem[Wan et~al.(2025)]{wan2025loongflow}
Chunhui Wan et~al.
\newblock {LoongFlow}: Directed evolutionary search via a cognitive
  plan-execute-summarize paradigm, 2025.
\newblock arXiv:2512.24077.

\bibitem[Yue et~al.(2025)Yue, Chen, Lu, Zhao, Wang, Yue, Song, and
  Huang]{yue2025rlvr}
Yang Yue, Zhiqi Chen, Rui Lu, Andrew Zhao, Zhaokai Wang, Yang Yue, Shiji Song,
  and Gao Huang.
\newblock Does reinforcement learning really incentivize reasoning capacity in
  {LLMs} beyond the base model?
\newblock In \emph{Advances in Neural Information Processing Systems}, 2025.
\newblock arXiv:2504.13837.

\bibitem[Zeigarnik(1927)]{zeigarnik1927behalten}
Bluma Zeigarnik.
\newblock {Das Behalten erledigter und unerledigter Handlungen}.
\newblock \emph{Psychologische Forschung}, 9:\penalty0 1--85, 1927.

\bibitem[Zelikman et~al.(2022)Zelikman, Wu, Mu, and Goodman]{zelikman2022star}
Eric Zelikman, Yuhuai Wu, Jesse Mu, and Noah~D. Goodman.
\newblock {STaR}: Bootstrapping reasoning with reasoning.
\newblock In \emph{Advances in Neural Information Processing Systems}, 2022.

\end{thebibliography}

\end{document}